# Noise Sensitivity of Local Descriptors vs ConvNets: An application to Facial Recognition

Y.M. Ayami
Durban University of Technology
Department of Information
Technology, Ritson Campus, Durban,
South Africa
ayamlearning@gmail.com

A. Shabat
Durban University of Technology
Department of Information
Technology, Ritson Campus, Durban,
South Africa
abshabat@gmail.com

D. Heukelman
ICTAS at Durban University of
Technology
Department of Information
Technology, Ritson Campus, Durban,
South Africa
deleneh@dut.ac.za

## ABSTRACT

The Local Binary Patterns (LBP) is a local descriptor proposed by Ojala et al to discriminate texture due to its discriminative power. However, the LBP is sensitive to noise and illumination changes. Consequently, several extensions to the LBP such as Median Binary Pattern (MBP) and methods such as Local Directional Pattern (LDP) have been proposed to address its drawbacks. Though studies by Zhou et al, suggest that the LDP exhibits poor performance in presence of random noise. Recently, convolution neural networks (ConvNets) were introduced which are increasingly becoming popular for feature extraction due to their discriminative power. This study aimed at evaluating the sensitivity of ResNet50, a ConvNet pre-trained model and local descriptors (LBP and LDP) to noise using the Extended Yale B face dataset with 5 different levels of noise added to the dataset. In our findings, it was observed that despite adding different levels of noise to the dataset, ResNet50 proved to be more robust than the local descriptors (LBP and LDP).

## 1 INTRODUCTION

The usage of devices with one or more cameras has become more widespread. These devices are mainly used for communication whilst others exploit them to capture images and subsequently, share their memories on social media platforms such as Facebook and Instagram. However, during the acquisition of images, these images are prone to noise. Image noise is the random variation of brightness or color information in images (Zaitoun and Aqel 2015). Consequently, image noise results in the obscuring of the desired information.

If an image is provided to a computer, the first step of the computer is semantic understanding which involves extracting efficient and effective features and build models based on these extracted features which makes the feature extraction process a crucial task and complicated task. Since the performance of our models is largely dependent on these extracted features. The most common features include color, texture and shape (Tang, Alelyani and Liu 2014). Local descriptors such as LBP and LDP have been exploited for feature extraction in real world applications such as medical image analysis, biometrics and security (Shabat and Tapamo 2016). Recently, convolution neural networks (ConvNets) were introduced which are increasingly becoming popular for feature extraction. Studies suggest that these models exhibit a strong discriminate power when discriminating texture. Arising from this, this study aimed to achieve the following:

a) Compare the sensitivity of LBP, LDP and ResNet50, a ConvNet pre-trained model to noise.
b) Compare the sensitivity of local descriptors (LBP and LDP) with varying parameters in the presence of noise.

It is worth mentioning that the Gaussian noise was applied to the dataset because it is more suited for photography images (Kylberg and Sintorn 2013) which are being considered in this study.

### 1.1 State of the Art

The Local Binary Pattern (LBP) is a local descriptor that is used to discriminate texture and was introduce by Ojala et al in the mid-90s as cited by (Nanni, Lumini and Brahnam 2012). The LBP is considered to have a low computational complexity and a high discriminative power. However, the LBP is sensitive to illumination variations and noise. Consequently, many extensions to the LBP descriptor have been proposed where the thresholding and encoding schemes are modified to create more robust descriptors (Kylberg and Sintorn 2013).

Kylberg and Sintorn (2013) evaluated the performance of eight LBP-based texture descriptors namely: Improved Local Binary Pattern (ILBP), Median Binary Pattern (MBP), Local Ternary Pattern (LTP), Improved Local Ternary Pattern (ILTP) , Robust Local Binary Patterns (RLBP), Fuzzy/soft Local Binary Patterns (FLBP), Shift Local Binary Patterns (SLBP) and Local Quinary Patterns (LQP) and compared them on six different datasets under increasing levels of addictive Gaussian white Noise together with Haralick descriptors, Gabor Filter and the classic LBP. It was established that the ILTP followed by FLBP produced outstanding results compared to the rest of the LBP-Family descriptors. The classic LBP was also outperformed in all tests performed.

Since the LBP performs poorly in the presence of illumination variations and noise, Jabid, Kabir and Chae (2010b) introduced the LDP to overcome these limitations. In order to achieve this, unlike LBP which considers surrounding neighboring pixels, LDP makes use of edge response values in all different directions. Subsequently, making the LDP to be robust in the presence of noise and illumination changes. The authors tested the performance of the local descriptors (LBP and LDP) on the FERET database. Based on the findings, it was established that LDP outperformed LBP. Raj et al. (2018) benchmarked the performance of handcrafted feature extraction techniques against ConvNets on AR and Extended Yale B face dataset. The findings proved that state of the art ConvNets models outperformed the traditional feature extraction-based technique (SIFT) in terms of recognition accuracy. However, it is worth noting that the former studies only considered clean images and the effect of noise on the features have not yet been studied. Which subsequently makes this study more relevant.

According to Zhou et al as cited by (Rivera, Castillo and Chae 2015), the LDP exhibits poor performance in presence of random noise. More so, Castillo, Rivera and Chae (2012) states that despite the LDP making use of edge directional information which makes its insensitive to noise and illumination changes, LDP lacks directional information since all the directions are treated equally. Consequently, making it sensitive to noise and illumination changes. Additionally, LDP makes use of edge response values in all different directions which subsequently makes it to be robust. However, during the generation of kirsch response values, the standard number of significant bits, $k$, has not been agreed upon in literature. Shabat and Tapamo (2016) are of the opinion that changing the value of $k$ has a significant effect on the performance of the LDP descriptor. This therefore forms the basis of this study.

Until recently, datasets of labeled images were relatively small, in the order of tens of thousands of images. Krizhevsky, Sutskever and Hinton (2012) asserts that simple recognition tasks such as facial recognition (Wechsler et al. 2012) and gender classification (Shan 2012) can be solved quite well with datasets of this size, using the traditional machine learning classifiers such as K-Nearest Neighbour (KNN), Logistic Regression (LR), Support Vector Machine (SVM), Multi-Layer Perceptron (MLP) and Naive Bayes (NB) especially if they are augmented with label-preserving transformations. The authors further argue that objects exhibit considerable variability, hence the need to use much larger training sets, as a result, the performance of recognition systems in an uncontrolled environment, such as variations in pose, noise, illumination, or expression still remain a challenge (DiCarlo, Pinto and Cox 2008; Huang et al. 2012).

Convolutional networks (ConvNets) have lately achieved great success in large-scale image and video recognition due to the availability of large public image repositories, such as ImageNet and high-performance computing systems, such as GPUs or large-scale distributed clusters (Simonyan and Zisserman 2014). Despite ConvNets achieving good results, these models require increased amounts of data and computational power such as GPUs. To overcome this limitation, pre-trained models, which are deep learning model weights that can be downloaded and used without training, were introduced. Examples of publicly available pre-trained models include: the VGG16, VGG19, Inception-v3, and ResNet50. These pre-trained models can be used for prediction, feature extraction, and fine-tuning.

## 1.2 Contribution
The contributions of this study are as follows:

- contribution to the existing scientific knowledge on the sensitivity of LBP, LDP and Resnet50 to noise;
- as well as the impact of optimizing parameters on the performance of LBP and LDP.

The rest of the study is categorized as follows: Section Two presents literature on the extraction methods used which include LBP, LDP and ResNet50. Section Three explains how the experiments were set up and the describes the parameters used. Section Four describes the findings and discussed these findings. Finally, Section Five concludes the paper and future work is proposed.

## 2 LOCAL FEATURES FOR TEXTURE ANALYSIS

### 2.1 Local Binary Pattern (LBP)

The LBP is a local texture descriptor which is considered to have of low computational complexity and discriminative. The LBP works by assigning a label to every pixel of an image by thresholding the 3 × 3 neighborhoods of each pixel with the center pixel value and considering the result as a binary number. A 256-bin histogram of LBP labels computed over the region is used as a texture descriptor (Shan, Gong and McOwan 2009). However, the LBP still has prominent limitations such as sensitivity to illumination changes and noise (Jabid, Kabir and Chae 2010b; Mohamed *et al.* 2014).

### 2.2 Local Directional Pattern (LDP)
Jabid et al Jabid, Kabir and Chae (2010a) proposed the LDP, a feature extraction method, to overcome the existing drawbacks of the LBP, such as illumination changes and noise (Jabid, Kabir and Chae 2010b). The LDP is based on the known Kirsch kernels. Unlike the LBP, the LDP has eight different directions wherein the edge response values are considered (Shabat and Tapamo 2014). The LDP features are composed of an eight-bit binary code. Each pixel of an input image is assigned to this code

### 2.3 ResNet50
Originally proposed by He *et al.* (2016), Residual Networks (ResNets) are deep convolutional neural networks that achieved good accuracy in ILSVRC 2015 ImageNet challenge with a Rank 1 classification task. These networks demonstrated a unique approach, where instead of learning unreferenced functions in the network, they explicitly reformulate the layers as learning residual functions with reference to the layer inputs. ResNet contains 5 types of configurations, each with different input/output dimensions, filter size, stride size, pooling size, and pooling stride size. These configurations are inclusive of ResNet18 with 18-layers; ResNet34 with 34-layers; ResNet50 with 50 layers; ResNet101 with 101-leayers and finally ResNet152 which has a total of 152-layers.

The input image to the network is resized to a fixed dimension of (224 × 224) pixels. ResNets perform all the standard operations of a ConvNets such as convolution, max-pooling, and batch normalization. After each convolution and before activation, batch normalization is performed. Stochastic Gradient Descent (SGD) is used as the optimizer with a batch size of 256. Based on the error that is getting accumulated, the learning rate is adjusted accordingly (from an initial value of 0.1), momentum is chosen as 0.9, and weight decay is chosen as 0.0001. The overall network is trained for 6,00,000 epochs. The ResNet50 pre-trained model was used for this study because it shows better performance in terms of accuracy and is computationally efficient compared to other pre-trained models (Islam *et al.* 2017)

## 3 EXPERIMENTAL SETUP

### 3.1 Data Set

This study made use of images from the publicly available extended Yale face database b, which contains a total of 16128 images of 28 human subjects under 9 poses and 64 illumination conditions (Georghiades, Belhumeur and Kriegman 2001; Lenc and Král 2015) . However, in this experiment, 10 human subjects under 9 poses and 64 illumination conditions were used. All images used were gray level images with size 100 × 100 pixels. The data was split into 20 % test and 80% training data.

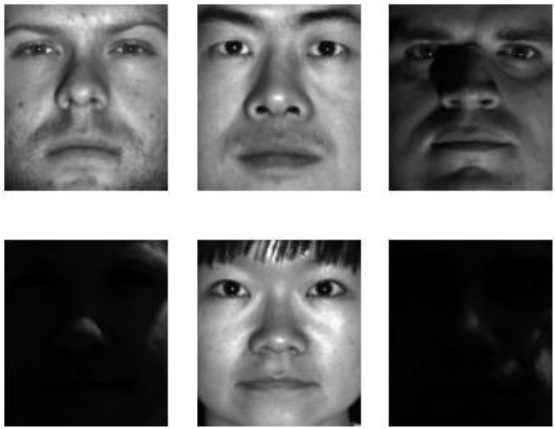

Figure 2: Sample face images from Extended Yale B face dataset

However, it is worth mentioning that the cropped version of the Yale face database b was used. Consequently, there was no need to crop the face from the images.

### 3.2 Parameter Optimization

The parameters for the local feature extraction methods (LBP and LDP) used in this study are summarized in the table below:

Table 1 Local Descriptor parameters and set used during parameter optimization

| Local Feature Extraction Method | Parameter | Set |
|---|---|---|
| LBP | Number of sample points (N) | $N \in \{8, 16\}$ |
| LBP | Radius (R) | $R \in \{1, 2\}$ |
| LDP | Number of Significant Bit (K) | $K \in \{3, 5\}$ |

### 3.3 Introducing Noise

The study then applied Gaussian noise to the original dataset as shown in Figure 3 and the noisy datasets were saved. The 4 noise levels, σ, used, include 0.0006, 0.007, 0.0785 and 0.8859.

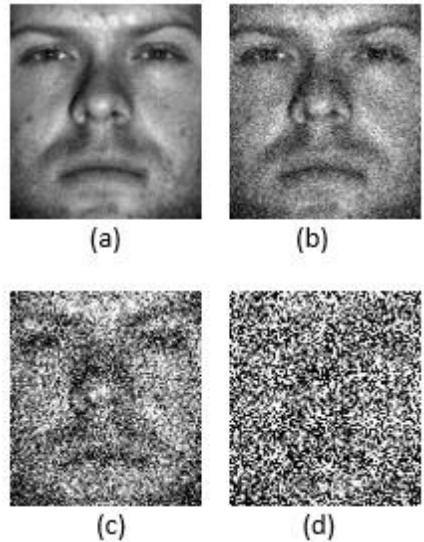

Figure 3: Sample face images with 4 different levels of noise. Image (a) shows the noise level at 0.0006 (b) shows the noise level at 0.007 (c) shows the noise level at 0.0785 and (d) shows the noise level at 0.8859.

## 4 FINDINGS AND DISCUSSION

### 4.1 Parameter Optimization for Local Descriptors

Table 1 lists the parameter values used during the optimization process of local descriptors (LBP and LDP). Our findings confirmed the assertions of Shabat and Tapamo (2016) that optimizing the parameters had a significant impact on the performance of local descriptors (LBP and LDP). For example, in Table 2, when the value of *R* for LBP was 1 and *N* was 8, the accuracy was at 74.22 % using the NB classifier. After optimizing the parameters ($R = 2$ and $N = 16$), the performance subsequently increased to 92.19%. The same was observed for the LDP were optimizing the parameters had an impact on the accuracy. Overall, the LDP performed better with the value of $k = 3$ where as the LBP performed better with the values $R = 2$ and $N = 16$.

## 4.2 Comparison of extraction methods without added noise

Table 2 depicts the findings of the extraction methods (LBP, LDP and ResNet) used in this study, in the absence of noise. From the findings, the mean accuracy shows that all the extraction methods performed well with the accuracy ranging between 66.09% to 88.91% LDP ($k = 3$) and ResNet had an outstanding accuracy of 100% using the LR classifier, whilst the LDP ($k = 5$) had the worst performance of 48.4% using the NB classifier. Surprisingly, the mean accuracy of LBP ($R = 2, N = 16$) which was 91.8% outperformed the LDP as well as ResNet. It is worth mentioning that the LR classifier achieved remarkable results, whilst NB in most cases had the worst performance.

## 4.3 Robustness to noise

In a landmark paper by Jabid, Kabir and Chae (2010a) it was established that LDP outperformed LBP in the presence of noise and illumination changes. In this study, these findings were benchmarked, using the Yale face database b with various levels of noise added to the dataset. Tables 2,3,4,5 and 6 summarize our findings using four classifiers namely KNN, NB, LR and MLP. In consonance to the findings of Jabid et. (2010a), the LDP was less sensitive to noise compared to the LBP using the LR classifier when the noise level was at 0. But as the noise level gradually increased to 0.0006 and 0.007, it was observed that the difference in accuracy between the two local descriptors narrowed.

In line with the findings of Raj *et al.* (2018), after comparing the mean accuracies of all extraction methods on the extended Yale face database b with various levels of noise applied to the dataset as can be observed in Tables 2,3,4,5 and 6, LBP and LDP were outperformed by ResNet50. It can therefore be concluded that ResNet50 is less sensitive to noise and illuminations changes compared to LBP and LDP. Furthermore, it can be concluded that the LDP is less sensitive to noise and illumination changes than LBP when the noise levels are less. However, as the noise levels increase, the sensitivity of the LDP increases compared to that of LBP. And subsequently, the LBP outperforms the LDP.

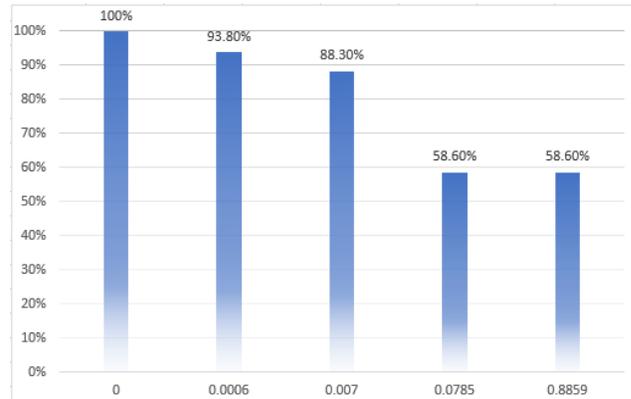

Figure 4: Highest accuracy for each noise level.

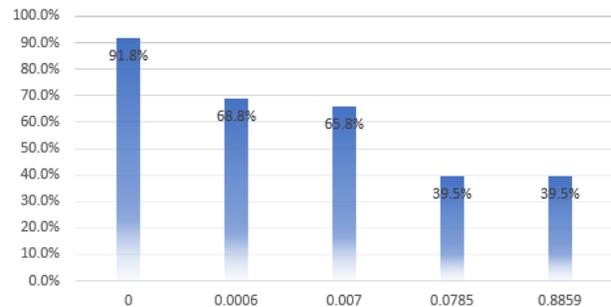

Figure 5: Highest mean accuracy for each noise level.

Table 2 classification accuracy for LBP, LDP and Resnet50 with on the dataset without noise applied to the dataset

| Feature Extraction Method | KNN (%) | NB (%) | LR (%) | MLP (%) | Mean Value (%) | Std Dev Value (%) |
|---|---|---|---|---|---|---|
| LBP (R=1, N=8) | 78.1 | *74.2* | **86.7** | 85.9 | 81.2 | 6.1 |
| LBP (R=2, N=16) | *89.9* | 92.2 | **93.0** | 92.2 | **91.8** | 1.3 |
| LDP (k=3) | 90.6 | *57.0* | **100.0** | 87.5 | 83.8 | 18.6 |
| LDP (k=5) | 82.8 | *48.4* | **93.0** | 75.0 | 74.8 | 19.1 |
| ResNet50 | 86.7 | *71.9* | **100.0** | 98.4 | 89.3 | 13.0 |

**\*For each feature extraction method, Bold results represent the highest accuracy; Italic results the least accuracy and the highlighted result represents the overall highest accuracy.**

Table 3 classification accuracy for LBP, LDP and Resnet50 with on the dataset with noise level ($\theta$) 0.0006 applied to the dataset

| Feature Extraction Method | KNN (%) | NB (%) | LR (%) | MLP (%) | Mean Value (%) | Std Dev Value (%) |
|---|---|---|---|---|---|---|
| LBP (R=1, N=8) | 66.4 | 37.5 | 53.1 | **68.0** | 56.3 | 14.2 |
| LBP (R=2, N=16) | 72.7 | 53.9 | 66.4 | **82.0** | **68.8** | 11.8 |
| LDP (k=3) | 79.7 | 40.6 | **92.2** | 75.8 | 72.1 | 22.1 |
| LDP(k=5) | 61.7 | *28.9* | **82.0** | 38.3 | 52.7 | 23.9 |
| ResNet50 | 69.5 | *66.4* | **93.8** | 77.3 | 56.3 | 14.2 |

* **For each feature extraction method, Bold results represent the highest accuracy; Italic results the least accuracy and the highlighted result represents the overall highest accuracy.**

Table 4 classification accuracy for LBP, LDP and Resnet50 with on the dataset with noise level ($\theta$) 0.007 applied to the dataset

| Feature Extraction Method | KNN (%) | NB (%) | LR (%) | MLP (%) | Mean Value (%) | Std Dev Value (%) |
|---|---|---|---|---|---|---|
| LBP (R=1, N=8) | 38.3 | 28.9 | *19.5* | **43.8** | 32.6 | 10.7 |
| LBP (R=2, N=16) | 53.9 | 50.8 | *35.2* | **66.4** | 51.6 | 12.8 |
| LDP (k=3) | 46.1 | *27.3* | **68.8** | 38.3 | 45.1 | 17.6 |
| LDP(k=5) | 27.3 | *17.2* | **49.2** | 18.8 | 28.1 | 14.7 |
| ResNet50 | *54.7* | 58.6 | **88.3** | 61.7 | **65.8** | 15.3 |

* **For each feature extraction method, Bold results represent the highest accuracy; Italic results the least accuracy and the highlighted result represents the overall highest accuracy.**

Table 5 classification accuracy for LBP, LDP and Resnet50 with on the dataset with noise level ($\theta$) 0.0785 applied to the dataset

| Feature Extraction Method | KNN (%) | NB (%) | LR (%) | MLP (%) | Mean Value (%) | Std Dev Value (%) |
|---|---|---|---|---|---|---|
| LBP (R=1, N=8) | 25.8 | 21.1 | *16.4* | **40.6** | 26.0 | 10.5 |
| LBP (R=2, N=16) | 31.3 | *21.9* | 23.4 | **52.3** | 32.2 | 14.0 |
| LDP (k=3) | 18.8 | *14.1* | **21.9** | 21.1 | 19.0 | 3.5 |
| LDP(k=5) | 25.0 | *13.3* | **32.8** | 24.2 | 23.8 | 8.0 |
| ResNet | 36.7 | *16.4* | **58.6** | 46.1 | **39.5** | 17.8 |

* **For each feature extraction method, Bold results represent the highest accuracy; Italic results the least accuracy and the highlighted result represents the overall highest accuracy.**

Table 6 classification accuracy for LBP, LDP and Resnet50 with on the dataset with noise level (σ) 0.8859 applied to the dataset

| Feature Extraction Method | KNN (%) | NB (%) | LR (%) | MLP (%) | Mean Value (%) | Std Dev Value (%) |
|---|---|---|---|---|---|---|
| LBP (R=1, N=8) | 21.9 | 15.6 | *10.9* | **25.8** | 18.6 | 6.6 |
| LBP (R=2, N=16) | 23.4 | 18.0 | *12.5* | **32.0** | 21.5 | 8.3 |
| LDP (k=3) | **18.0** | *7.8* | 17.2 | 11.7 | 13.7 | 4.8 |
| LDP(k=5) | 13.3 | *7.8* | 12.5 | **18.0** | 12.9 | 4.2 |
| ResNet | 36.7 | *16.4* | 58.6 | 46.1 | **39.5** | 17.8 |

\* **For each feature extraction method, Bold results represent the highest accuracy; Italic results the least accuracy and the highlighted result represents the overall highest accuracy.**

## 5 CONCLUSIONS AND FUTURE WORK

This study evaluated the sensitivity of ResNet50, a ConvNet Pre-Trained Model and local descriptors (LBP and LDP) in the presence of varying levels of Gaussian noise on the extended Yale b face database. From our findings, the following was observed:

- It was observed that despite increasing the levels of noise, ResNet50 was less sensitive to noise compared to LDP and LBP in all the five experiments conducted.

- When local descriptors (LBP and LDP) were compared, it was observed that LDP performs relatively well on noise free data. However, as the noise levels increased, LDP become more sensitive to noise than LBP.

- Adjusting parameters had a significant effect on the performance of the local descriptors.

Based on these findings, it is safe to say that ResNet50 pre-trained models are the most suitable models for feature extraction compared to LBP and LDP. However, several other ConvNet models exist such as VGG16, VGG19 and Inception-v3. Future work could focus on:

- Evaluating the sensitivity of pre-trained models to noise and analyse the sensitivity with respect to the local descriptors.
- Comparing the computational cost of pre-trained models to local descriptors during feature extraction.

## REFERENCES

Georghiades, A. S., Belhumeur, P. N. and Kriegman, D. J. 2001. From few to many: Illumination cone models for face recognition under variable lighting and pose. *IEEE Transactions on pattern analysis and machine intelligence*, 23 (6): 643-660.

He, K., Zhang, X., Ren, S. and Sun, J. 2016. Deep residual learning for image recognition. In: Proceedings of *Proceedings of the IEEE conference on computer vision and pattern recognition*. 770-778.

Islam, M. T., Aowal, M. A., Minhaz, A. T. and Ashraf, K. 2017. Abnormality detection and localization in chest x-rays using deep convolutional neural networks. *arXiv preprint arXiv:1705.09850*,

Jabid, T., Kabir, M. H. and Chae, O. 2010a. Local directional pattern (LDP) for face recognition. In: Proceedings of *Consumer Electronics (ICCE), 2010 Digest of Technical Papers International Conference on*. IEEE, 329-330.

Jabid, T., Kabir, M. H. and Chae, O. 2010b. Robust facial expression recognition based on local directional pattern. *ETRI journal*, 32 (5): 784-794.

Kylberg, G. and Sintorn, I.-M. 2013. Evaluation of noise robustness for local binary pattern descriptors in texture classification. *EURASIP Journal on Image and Video Processing*, 2013 (1): 17.

Lenc, L. and Král, P. 2015. Unconstrained Facial Images: Database for face recognition under real-world conditions. In: Proceedings of *Mexican International Conference on Artificial Intelligence*. Springer, 349-361.

Mohamed, M. A., Rashwan, H. A., Mertsching, B., García, M. A. and Puig, D. 2014. Illumination-robust optical flow using a local directional pattern. *IEEE Transactions on Circuits and Systems for Video Technology*, 24 (9): 1499-1508.

Nanni, L., Lumini, A. and Brahnam, S. 2012. Survey on LBP based texture descriptors for image classification. *Expert Systems with Applications*, 39 (3): 3634-3641.

Raj, M., Gogul, I., Raj, M. D., Kumar, V. S., Vaidehi, V. and Chakkaravarthy, S. S. 2018. Analyzing ConvNets Depth for Deep Face Recognition. In: Proceedings of *Proceedings of 2nd International Conference on Computer Vision & Image Processing*. Springer, 317-330.

Rivera, A. R., Castillo, J. R. and Chae, O. 2015. Local directional texture pattern image descriptor. *Pattern Recognition Letters*, 51: 94-100.

Shabat, A. M. and Tapamo, J.-R. 2016. Directional local binary pattern for texture analysis. In: Proceedings of *International Conference Image Analysis and Recognition*. Springer, 226-233.

Shabat, A. M. and Tapamo, J. R. 2014. A comparative study of Local directional pattern for texture classification. In: Proceedings of *Computer Applications & Research (WSCAR), 2014 World Symposium on*. IEEE, 1-7.

Shan, C., Gong, S. and McOwan, P. W. 2009. Facial expression recognition based on local binary patterns: A comprehensive study. *Image and Vision Computing*, 27 (6): 803-816.

Tang, J., Alelyani, S. and Liu, H. 2014. Feature selection for classification: A review. *Data Classification: Algorithms and Applications*: 37.

Zaitoun, N. M. and Aqel, M. J. 2015. Survey on image segmentation techniques. *Procedia Computer Science*, 65: 797-806.